\documentclass[sigconf]{acmart}

\usepackage{balance}

\graphicspath{{images/}}

\AtBeginDocument{%
  \providecommand\BibTeX{{%
    \normalfont B\kern-0.5em{\scshape i\kern-0.25em b}\kern-0.8em\TeX}}}

\begin{document}
\fancyhead{}

\title{\emph{WhatTheWikiFact}: Fact-Checking Claims Against Wikipedia}

\author{Anton Chernyavskiy}
\affiliation{%
  \institution{HSE University}
  \city{Moscow}
  \country{Russia}}
\email{aschernyavskiy\_1@edu.hse.ru}

\author{Dmitry Ilvovsky}
\affiliation{%
  \institution{HSE University}
  \city{Moscow}
  \country{Russia}}
\email{dilvovsky@hse.ru}

\author{Preslav Nakov}
\affiliation{%
  \institution{Qatar Computing Research Institute, HBKU}
  \city{Doha}
  \country{Qatar}}
\email{pnakov@hbku.edu.qa}



\begin{abstract}
  The rise of Internet has made it a major source of information. Unfortunately, not all information online is true, and thus a number of fact-checking initiatives have been launched, both manual and automatic, to deal with the problem. Here, we present our contribution in this regard: \emph{WhatTheWikiFact}, a system for automatic claim verification using Wikipedia. The system can predict the veracity of an input claim, and it further shows the evidence it has retrieved as part of the verification process. It shows confidence scores and a list of relevant Wikipedia articles, together with detailed information about each article, including the phrase used to retrieve it, the most relevant sentences extracted from it and their stance with respect to the input claim, as well as the associated probabilities. The system supports several languages: Bulgarian, English, and Russian.
\end{abstract}

\ccsdesc[500]{Information systems~Information retrieval}
\ccsdesc[300]{Computer systems organization~Real-time systems}
\ccsdesc[500]{Applied computing~Document management and text processing}
\ccsdesc[300]{Information systems~Information systems applications}
\ccsdesc[500]{Information systems~Web searching and information discovery}
\ccsdesc[500]{Computing methodologies~Natural language processing}
\ccsdesc[500]{Computing methodologies~Information extraction}
\keywords{Fact-checking, Factuality, Veracity, Fake News, Disinformation, Misinformation, Document Retrieval, Sentence Retrieval, Stance Detection, Natural Language Inference, Wikipedia.}


\maketitle

\section{Introduction}

Internet is abundant in platforms that allow users to share information online such as social networks, blogs, and forums. Unfortunately, not all information online is true, and there is a need to verify questionable claims that users encounter online.

As manual fact-checking is a complex and time-consuming task, it is important to develop tools that can help automate the process. Various task formulations have been proposed to automate fact-checking, e.g.,~SemEval RumourEval tasks \cite{derczynski-etal-2017-semeval, gorrell-etal-2019-semeval}, the SemEval task on Fact Checking in Community Question Answering Forums \cite{mihaylova-etal-2019-semeval}, the CLEF CheckThat! lab \cite{clef2018checkthat:overall,CheckThat:ECIR2019,CheckThat:ECIR2020,CheckThat:ECIR2021}, the Fake News Challenge \cite{hanselowski-etal-2018-retrospective}, and the FEVER task on Fact Extraction and VERification \cite{thorne-etal-2018-fact,thorne-etal-2019-fever2,aly2021feverous}. Here, we focus on the FEVER task, as it offers a large-scale training dataset, and enables explainable systems.

Interestingly, despite the popularity of research using the FEVER task formulation and its dataset, to the best of our knowledge, there are no publicly available running systems based on it. Here, we aim to bridge this gap with our \emph{WhatTheWikiFact} system,\footnote{The \emph{WhatTheWikiFact} system is running online at\\ \url{https://www.tanbih.org/whatthewikifact}} which allows users to check claims against Wikipedia. It supports several languages (Bulgarian, English, and Russian) and uses a local Wikipedia for each language. The system first identifies relevant Wikipedia pages, and then finds relevant sentences within them. Then, it analyzes each claim submitted as an input against all extracted text fragments and makes a final verdict on the veracity of the claim: \texttt{Truth}, \texttt{Lying}, or \texttt{Not Enough Info}. \emph{WhatTheWikiFact} further displays all the information it used to make its decision, together with intermediate results for each verified claim, which includes (\emph{i})~a list of the titles of the most relevant documents with links to the corresponding Wikipedia pages, and (\emph{ii})~detailed information about each document as a table of retrieved text fragments $\langle$\texttt{position in the document | text fragment}$\rangle$, and a bar chart showing the confidence of the classifier in each label (\texttt{Supports}, \texttt{Refutes} or \texttt{Not Enough Info}) for each text fragment. This allows the user to quickly analyze the result.

The remainder of this paper is organized as follows: Section~\ref{sec:related} discusses related work. Section~\ref{sec:data} describes the dataset we used for training. Section~\ref{sec:system} offers an overview of the system and its components. Section~\ref{sec:implementation} discusses the core implementation details. Section~\ref{sec:evaluation} presents some evaluation results.  Section~\ref{sec:interface} describes the system interface and its functionality, with some examples. Finally, Section~\ref{sec:conclusion} concludes and discusses future work.

\section{Related Work}
\label{sec:related}

Many task formulations have been proposed to address the spread of misinformation and disinformation online, and for each formulation, a number of approaches have been tried. Some good readings on the topic include surveys such as that by \citet{Shu:2017:FND:3137597.3137600}, who adopted a data mining perspective on ``fake news'' and focused on social media.
Another survey \cite{10.1145/3161603} studied rumor detection in social media. 
The survey by \citet{thorne-vlachos-2018-automated} took a fact-checking perspective on ``fake news'' and related problems.

\citet{Li:2016:STD:2897350.2897352} covered truth discovery in general.
\citet{Lazer1094} offered a general overview and discussion on the science of ``fake news'', while
\citet{Vosoughi1146} focused on the process of proliferation of true and false news online.
Other recent surveys focused on stance detection~\cite{10.1145/3369026}, propaganda~\cite{ijcai2020-672}, social bots~\cite{10.1145/2818717}, false information~\cite{DBLP:journals/jdiq/ZannettouSBK19} and bias on the Web \cite{Baeza-Yates:2018:BW:3229066.3209581}. Some very recent surveys focused on stance for misinformation and disinformation detection \cite{Survey:2021:Stance:Disinformation}, on automatic fact-checking to assist human fact-checkers \cite{Survey:2021:AI:Fact-Checkers}, on predicting the factuality and the bias of entire news outlets \cite{Survey:2021:Media:Factuality:Bias}, and on multimodal disinformation detection \cite{Survey:2021:Multimodal:Disinformation}.

\paragraph{Non-explainable fact-checking} The primary focus of our system and of this paper is fact-checking of claims. Relevant research includes work on credibility assessment in Twitter, which has been addressed using user-based, message-based, topic-based, and propagation-based features \cite{10.1145/1963405.1963500}. \citet{rashkin-etal-2017-truth} analyzed the linguistic features used in the claims. \citet{wang-2017-liar} presented the LIAR dataset, which focuses on fact-checking using only the input claim (its text and metadata). \citet{lee-etal-2021-towards} found that misinformation can be discovered using perplexity analysis of the input claim, as perplexity is higher for false claims. A number of studies were also conducted on the feasibility of using language models for open-domain question answering and further as fact-checkers \cite{petroni-etal-2019-language, roberts-etal-2020-much, lee-etal-2020-language}. 
All this work was non-explainable. 

\paragraph{Explainable fact-checking} This is a more relevant direction. \citet{shaar-etal-2020-known} developed two datasets for detecting previously fact-checked claims, which were extended and used for shared tasks as part of the CLEF CheckThat! lab in 2020 and 2021 \cite{clef-checkthat:2020,clef-checkthat-ar:2020,clef-checkthat-en:2020,CheckThat:ECIR2020,clef-checkthat:2021:task2,clef-checkthat:2021:LNCS,CheckThat:ECIR2021}. \citet{DBLP:conf/iclr/ChenWCZWLZW20} proposed table-based fact verification, and \citet{10.1145/3308558.3314126} used knowledge graphs. 

\paragraph{Stance detection as an element of fact-checking} This was the objective of the Fake News Challenge task \cite{hanselowski-etal-2018-retrospective, Riedel2017ASB}, as well as of the RumourEval tasks at SemEval in 2017 and 2019 \cite{derczynski-etal-2017-semeval, gorrell-etal-2019-semeval}.

\paragraph{Fact-Checking Using Wikipedia} In our system, we use the FEVER dataset and task formulation, which enables Wikipedia-based explainable fact-checking
\cite{thorne-etal-2018-fever}. The dataset was used in the FEVER shared tasks \cite{thorne-etal-2018-fact,thorne-etal-2019-fever2}, where most systems had the following components: (\emph{i})~document retrieval, (\emph{ii})~sentence retrieval, and (\emph{iii})~natural language inference (NLI). \citet{otto-2018-team, chakrabarty-etal-2018-robust, hanselowski-etal-2018-ukp, alonso-reina-etal-2019-team} used a search API to retrieve relevant documents, while \citet{yoneda-etal-2018-ucl} used logistic regression. Word Mover's Distance \cite{chakrabarty-etal-2018-robust}, TF.IDF \cite{malon-2018-team}, ESIM \cite{hanselowski-etal-2018-ukp}, logistic regression \cite{yoneda-etal-2018-ucl}, BERT \cite{stammbach-neumann-2019-team} were used for sentence retrieval; and DAM \cite{otto-2018-team}, ESIM \cite{hidey-diab-2018-team}, Random Forest \cite{reddy-etal-2018-defactonlp}, LSTM \cite{Nie2019CombiningFE} and BERT \cite{stammbach-neumann-2019-team} were used for NLI. Here, we adopt a similar overall architecture.

\paragraph{System Demonstrations} Relevant demos include \emph{Hoaxy} \cite{10.1145/2872518.2890098}, for tracking misinformation in social networks and news sites, \emph{CredEye}, \cite{10.1145/3184558.3186967} for credibility assessment, \emph{Tracy} \cite{10.1145/3308558.3314126} for fact-checking using rules and knowledge graphs, 
\emph{Scrutinizer} \cite{10.14778/3407790.3407841} for fact-checking statistical claims,
\emph{STANCY} \cite{popat-etal-2019-stancy} for stance detection using BERT and consistency constraints, 
\emph{Tanbih} \cite{zhang-etal-2019-tanbih}, which analyzes news articles and media outlets and predicts factuality of reporting, degree of propaganda, hyper-partisanship, political bias, framing, and stance with respect to various claims and topics,
and \emph{FAKTA} \cite{nadeem-etal-2019-fakta} for stance and evidence extraction from the Web.

Unlike these systems, we perform textual evidence-based fact-checking using Wikipedia, following the FEVER formulation of the task. The most related demo, FAKTA, is also trained on data from FEVER, but it focuses on the stance of a document, e.g.,~retrieved from the Web, with respect to the input claim, while our system makes a prediction about a claim's factuality and gives sentence-level evidence from Wikipedia to explain its decision. Moreover, our system supports several languages besides English.

\section{Data}
\label{sec:data}

\begin{figure}[h]
\centering
\includegraphics[width=0.6\linewidth]{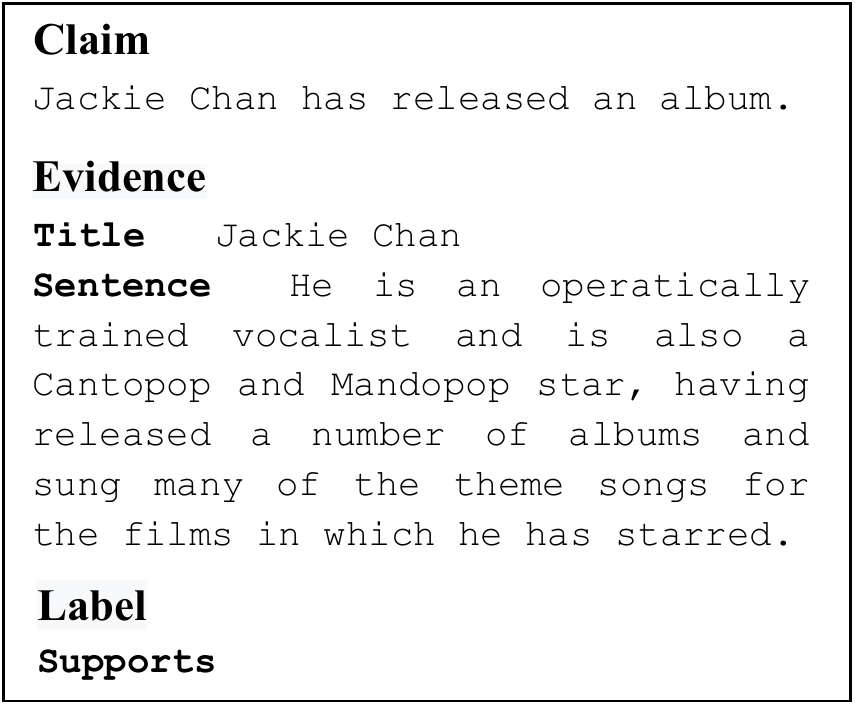}
\caption{Manually labeled claim from FEVER.}
\label{fever_example}
\end{figure}

We train our system on the FEVER dataset \cite{thorne-etal-2018-fever}, which includes a dump of 5.4M Wikipedia pages, and 220K claims labeled with one of the following three classes: \texttt{Supports}, \texttt{Refutes} or \texttt{Not Enough Info}. An example is shown in Figure \ref{fever_example}. For the former two labels, there is also evidence provided, i.e.,~a sentence or a set of sentences that would allow one to make a verdict about the veracity of the input claim, while the \texttt{Not Enough Info} label indicates that there is no enough evidence in Wikipedia to prove or to refute the claim. There can be different sets of evidence for a given claim, but all of them would support the same label.

\section{System Overview}
\label{sec:system}

\begin{figure}[h]
\centering
\includegraphics[width=1.0\linewidth]{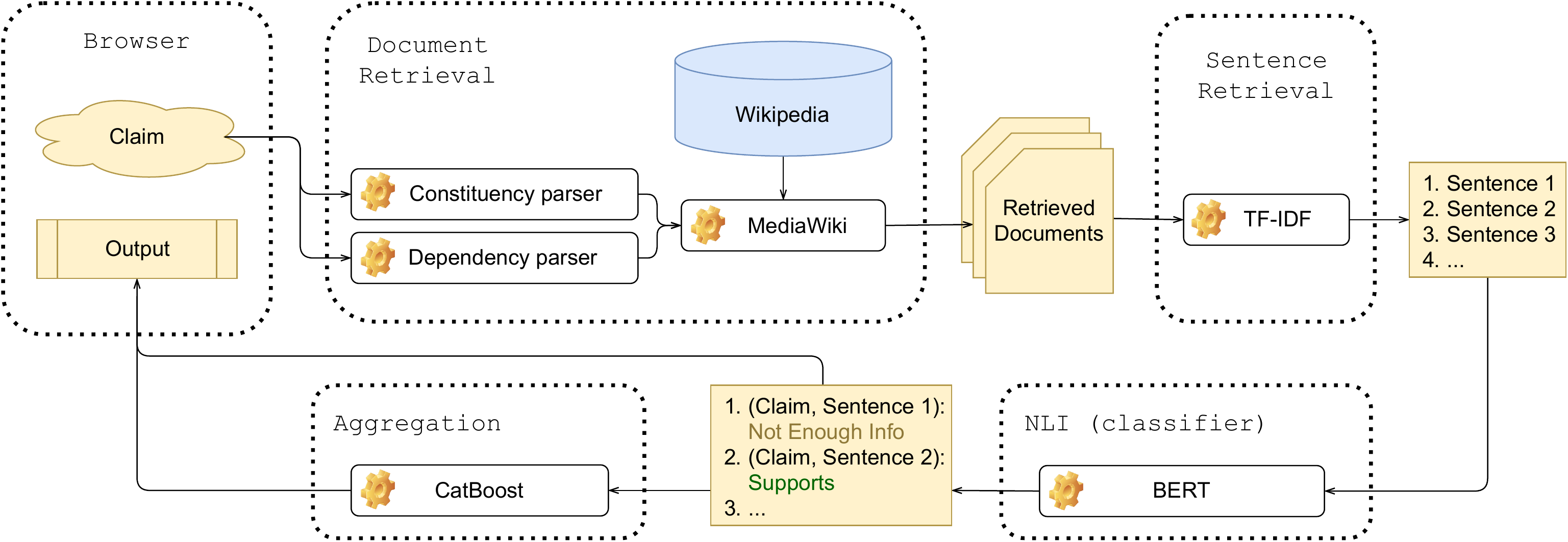}
\caption{The architecture of our \emph{WhatTheWikiFact} system.}
\label{pipeline}
\end{figure}

Figure \ref{pipeline} shows the general architecture of our system, which is similar to the one described in \cite{chernyavskiy-ilvovsky-2019-extract}. First, the Document Retrieval (DR) module finds potentially relevant documents from Wikipedia. Then, the Sentence Retrieval (SR) module extracts the top-20 most relevant sentences from these documents. Afterwards, the Natural Language Inference (NLI) module classifies each claim--sentence pair as support/refute/NEI. Finally, the aggregation module makes a final prediction. We describe these steps below in more detail.

\paragraph{Document Retrieval (DR)}

We use the Python MediaWiki API\footnote{\url{http://wikipedia.readthedocs.io/en/latest/}} to retrieve relevant documents from Wikipedia. We call the API with the results of our query generation module, which uses a constituency parser \cite{joshi-etal-2018-extending} to extract noun phrases, which we use as separate queries. We further generate a query from the part of the claim up to the head word, which we extract using a dependency parser \cite{DBLP:conf/iclr/DozatM17}. For each query, we retain the top-3 returned documents. As there can be many queries for an input claim, we further filter the results to improve the inference speed. Thus, we process the input claim and all retrieved titles shortened to the first bracket symbol using the Porter Stemmer \cite{Porter1980AnAF}, and we select them for the final set only if they are fully contained in the input.

\paragraph{Sentence Retrieval (SR)} 

This component selects the top-20 most relevant sentences from the documents retrieved by DR component. The relevance is estimated as the cosine between the TF.IDF representations of the claim and of the document titles. We apply a variant of TF.IDF that uses binarization of the non-zero term counts, and considers words with a document fraction higher than 0.85 as stopwords. Note that there are no thresholds for the relevance score, and thus all retrieved sentences can potentially have a score of zero, e.g.,~if all top-20 documents happen to be irrelevant.

\paragraph{Natural Language Inference (NLI)}

This component takes as input the source claim, a separator, a sentence retrieved by the SR component, another separator, and the title of the document. Here, we use all retrieved sentences independently. 
For this component, we use BERT \cite{devlin-etal-2019-bert}, which we fine-tune using a balanced training dataset. In particular, we use instances of the classes \texttt{Supports} and \texttt{Refutes} from the training set, and also the top-3 retrieved sentences for each claim, which we label as \texttt{Not Enough Info}. 

\paragraph{Aggregation}

We use CatBoost gradient boosting \cite{10.5555/3327757.3327770} to aggregate the sentence-level predictions of the NLI component. We train the model on part of the validation set using the stacked label probabilities predicted by the NLI model. Thus, we use a 60-dimensional (3 scores for 20 sentences) feature vector, potentially padded by zeros in case of lack of sentences in the retrieval phase.

We further report the maximal probability of the \texttt{Supports} (\texttt{Refutes}) class among all sentences scaled by the percentage of such sentences among the possibly relevant ones, i.e.,~excluding \texttt{Not Enough Info}, as the confidence score for \texttt{Truth} (\texttt{Lying}) predictions and the minimal probability of the \texttt{Not Enough Info}, otherwise.

\paragraph{Overall Architecture}
\label{sec:implementation}

\emph{WhatTheWikiFact} has a server and a client parts, which are connected via a REST API. The client part is implemented using the Streamlit framework\footnote{\url{http://streamlit.io/}}---it is the GUI part, that is, the interface in the browser. The server part uses the Fast API\footnote{\url{http://fastapi.tiangolo.com/}}. We use Allen NLP library \cite{gardner-etal-2018-allennlp} for the constituency and the dependency parsers for texts in English, and Natasha\footnote{\url{http://github.com/natasha/natasha}} and Stanza\footnote{\url{http://stanfordnlp.github.io/stanza/}} libraries for texts in Russian and Bulgarian, respectively. We use the official repository for BERT.\footnote{\url{http://github.com/google-research/bert}} We preload these models on the server, and we serve them using POST requests made by the client. 

The parsers receive a piece of text as input, which is then tokenized and stemmed using the NLTK library \cite{bird-loper-2004-nltk} for English and Russian, and using BulStem \cite{Nakov2003BulStemDA,Nakov2003:BulStem:Building} for Bulgarian. BERT receives a list of sentence pairs, preprocessed using the BERT tokenizer. 

Our NLI component is English-only, and thus we use the Google Translator API\footnote{\url{https://github.com/nidhaloff/deep-translator}} to translate the input into English first.

\section{Evaluation}
\label{sec:evaluation}

The machine learning model we use in our \emph{WhatTheWikiFact} system achieves an accuracy of 73.22 and a FEVER score of 67.44 on the test part of the FEVER dataset. Note that these scores are better than those for the best system at the FEVER shared task \cite{thorne-etal-2018-fact}, which had an accuracy of 68.21, and a FEVER score of 64.21. They are also on par with the best system from the builder phase of the FEVER2.0 shared task \cite{thorne-etal-2019-fever2}, which had a FEVER score of 68.46. We should note that there have been some better results reported in the literature since then. In fact, we also had stronger results in our offline experiments. However, we use this particular model, as we need real-time execution, which requires certain compromise in terms of accuracy for the sake of improved speed of execution, which is essential for a real-time system like ours.

Our analysis shows that the biggest fraction of the system classification errors are in distinguishing \texttt{Not Enough Info} sentences from the rest. At the same time, the accuracy of the retrieval component is almost 91\%, which means that in 91\% of the cases, we do retrieve correct evidence for the final set of potentially relevant sentences. Therefore, as the system classification errors are not very frequent, such cases can be easily analyzed by the user in our system's output.

\section{User Interface}
\label{sec:interface}

\begin{figure*}[ht]
\centering
\includegraphics[width=0.8\textwidth]{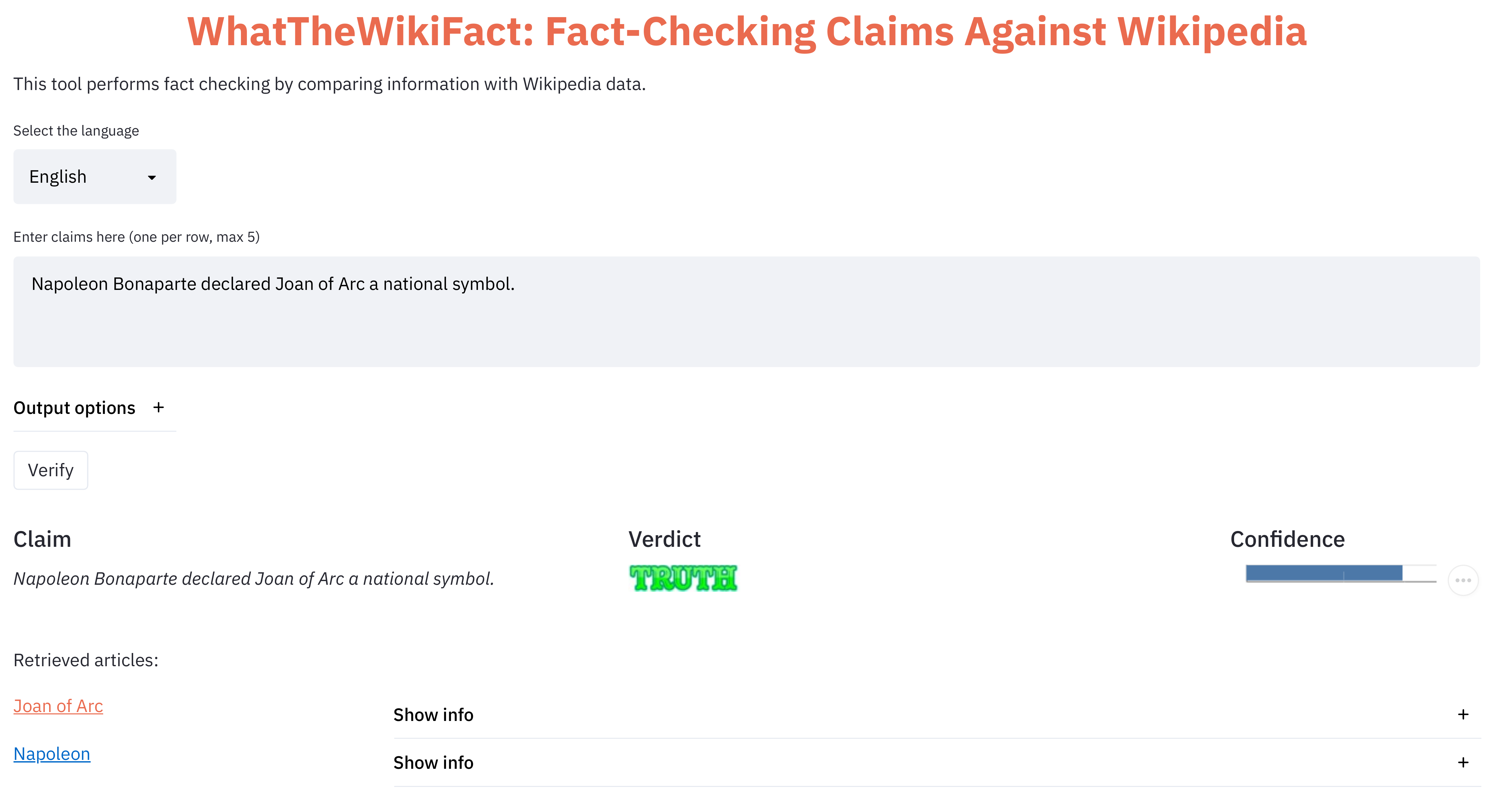}
\caption{Screenshot of what our \emph{WhatTheWikiFact} system outputs when verifying the claim: ``\emph{Napoleon Bonaparte declared Joan of Arc a national symbol.}''}
\label{example}
\end{figure*}

Figure \ref{example} shows a snapshot of \emph{WhatTheWikiFact}'s output for an input claim. We can see that the system offers an overview of the verification results, which includes a verdict, the system's confidence in that verdict, and a list of possibly relevant documents: title and a link to the corresponding Wikipedia page.

Figure \ref{example} shows that the system has retrieved three articles for the input claim ``\emph{Napoleon Bonaparte declared Joan of Arc a national symbol.}'' However, this does not mean that the system considers all of these articles as relevant. It just means that they contain some the top-20 most relevant sentences.

\begin{figure}[ht]
\includegraphics[width=1.0\linewidth]{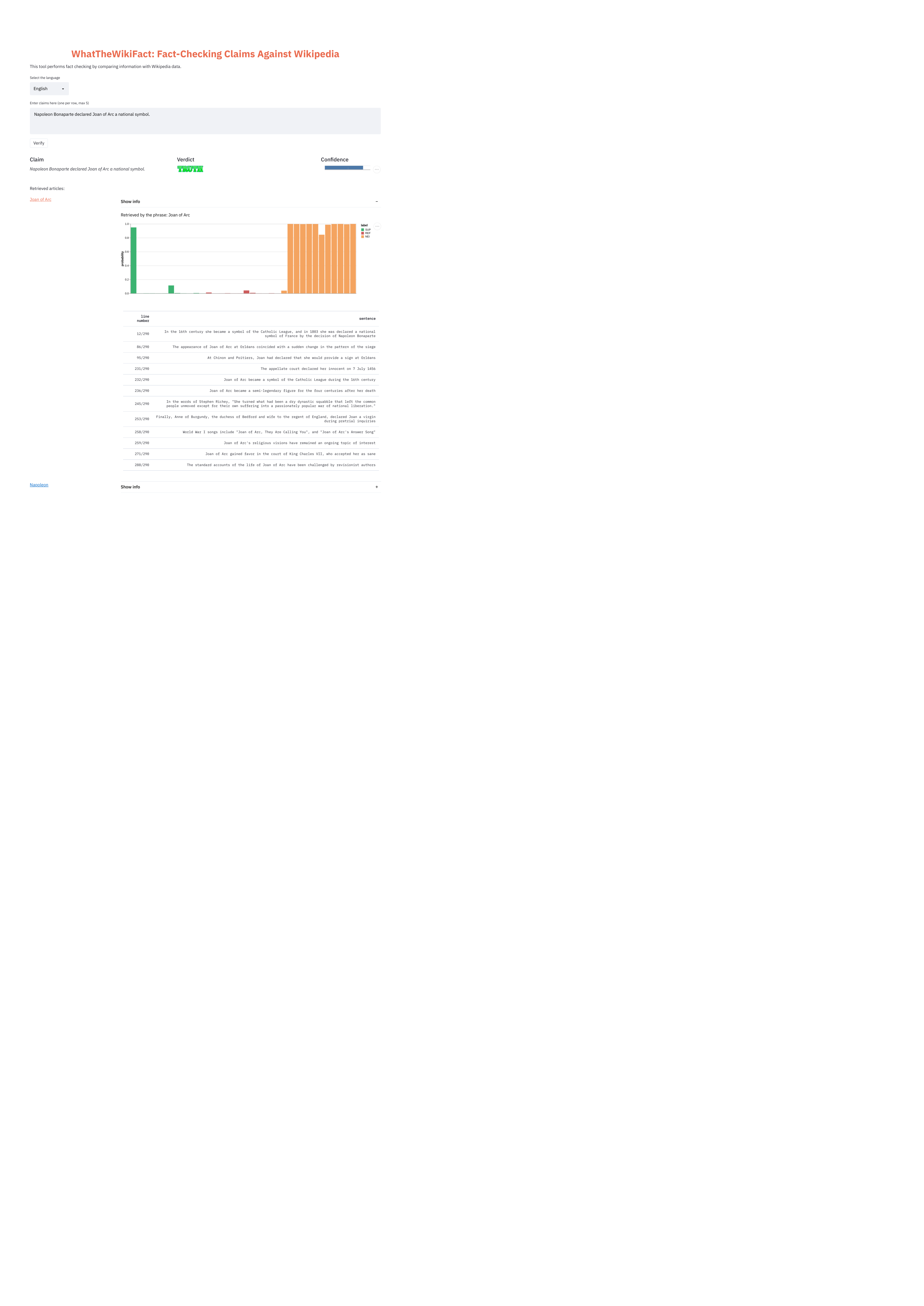}
\caption{Screenshot showing details about ``\emph{Joan of Arc}'', a Wikipedia article retrieved when verifying the claim ``\emph{Napoleon Bonaparte declared Joan of Arc a national symbol.}''}
\label{detailed}
\end{figure}

By clicking on \emph{Show info}, the user can expand the panel with details about each document. As a result, the following information will be shown (illustrated on Figure \ref{detailed}):

\begin{itemize}
    \item \textbf{The part of the input claim} used to retrieve the document.
    \item \textbf{A bar chart of the predicted stance labels for the input claim with respect to each retrieved sentence.} The stance is expressed as one of the classes \texttt{Supports} (SUP), \texttt{Refutes} (REF), or \texttt{Not Enough Info} (NEI). The chart further shows the class probability, which is also represented as the bar height, sentence number, and label, which is also indicated with the corresponding color. Note that there are three bars for each sentence, i.e., one for each label. Moreover, the bars are ordered (grouped) by labels or optionally by sentences (it is specified in the \emph{Output options} section).
    \item \textbf{A table of the most relevant sentences.} For each sentence in that table, we show its position in the document as well as the document length, thus reflecting also the relative position and allowing for easy matching with the bar chart.
\end{itemize}

For example, Figure \ref{detailed} shows that the article ``Joan of Arc'' is relevant for fact-checking the input claim, as it includes a sentence that supports it. Other retrieved sentences in this document have almost 100\% probability of a \texttt{Not Enough Info} label, and are thus irrelevant. Manual analysis by the user for the remaining two documents---``Napoleon'', which was retrieved by the phrase ``Napoleon Bonaparte'', and ``National symbol'', which was retrieved by the phrase ``a national symbol''---could confirm that they are indeed irrelevant. With these retrieval results, the user can manually inspect the boundary classification cases, especially for veracity prediction with low confidence or in case of \texttt{Not Enough Info}.

\section{Conclusion and Future Work}
\label{sec:conclusion}

We have presented \emph{WhatTheWikiFact}, a system for automatic claim verification using Wikipedia. The system reports the veracity for each input claim supplemented with evidence retrieved during the verification process, thus offering explainability. It also shows confidence scores and a set of relevant Wikipedia articles. Moreover, it allows the user to obtain detailed information about each article, including the exact phrase that was used to retrieve it, a list of the most relevant sentences with to the input claim that it contains, and their stance probabilities regarding the input claim.  The system supports several languages: Bulgarian, English, and Russian.

In future work, we plan to implement a more accurate model by distilling knowledge from a larger model. Another direction we want to explore is to add additional languages for verification using local Wikipedias, which can be implemented without additional training, e.g.,~using multilingual BERT or language adapters \cite{pfeiffer-etal-2020-adapterhub}. 

\section*{Acknowledgments}
This research was done within the framework of the HSE University Basic Research Program.

It is also part of the Tanbih mega-project (\url{http://tanbih.qcri.org/}), which is developed at the Qatar Computing Research Institute, HBKU, and aims to limit the impact of ``fake news,'' propaganda, and media bias by making users aware of what they are reading.

\bibliographystyle{ACM-Reference-Format}
\balance


\end{document}